# GAN-based Deidentification of Drivers' Face Videos: An Assessment of Human Factors Implications in NDS Data

Surendrabikram Thapa
Department of Computer Science, Virginia Tech
Blacksburg, USA
surendrabikram@vt.edu

Abhijit Sarkar
Virginia Tech Transportation Institute
Blacksburg, USA
asarkar@vtti.vt.edu

*Abstract*—This paper addresses the problem of sharing drivers' face videos for transportation research while adhering to proper ethical guidelines. The paper first gives an overview of the multitude of problems associated with sharing such data and then proposes a framework on how artificial intelligence-based techniques, specifically face swapping, can be used for de-identifying drivers' faces. Through extensive experimentation with an Oak Ridge National Laboratory (ORNL) dataset, we demonstrate the effectiveness of face-swapping algorithms in preserving essential attributes related to human factors research, including eye movements, head movements, and mouth movements. The efficacy of the framework was also tested on various naturalistic driving study data collected at the Virginia Tech Transportation Institute. The results achieved through the proposed techniques were evaluated qualitatively and quantitatively using various metrics. Finally, we discuss possible measures for sharing the de-identified videos with the greater research community.

*Index Terms*—Data Sharing, Privacy Protection, Human Factors, Face-swapping Algorithms, Equitable Transportation

## I. INTRODUCTION

Data sharing is crucial for collaborative research as it enables researchers to work together more efficiently and effectively, leading to new insights and discoveries that would not have been possible without data sharing. Another major benefit of data sharing is the ability to optimize resources by minimizing the re-collection of data [1]. However, when it comes to research involving human subjects, data sharing raises important ethical and privacy concerns that must be addressed. When data originating from human subjects are publicly shared, there are no guarantees that the data will always be used for legitimate research [2]. Improper use and access to data containing personally identifiable information (PII) violate the promises made to protect the privacy of those who consented to participate in the research. Therefore, it is important to develop de-identification techniques or privacy-focused data-sharing methods to ensure responsible data sharing while advancing scientific research.

Naturalistic driving studies (NDS) have been instrumental in understanding the role of driver behavior in transportation safety and mobility [3]. Analysis of drivers' behaviors

This work has been supported through funding from National Surface Transportation Safety Center for Excellence (NSTSCE).

from their face videos can provide valuable insights into the secondary behaviors (e.g., cell phone use), distraction, and drowsiness that can lead to safety-critical events. However, wide-scale distribution and use of face videos from NDS are restricted due to the presence of PII in the data. In order to eliminate such restrictions and enable broader research, de-identifying face videos from NDS is crucial. Face videos from NDS typically include two types of PII: basic informa- tion about the drivers, such as demographic information and physical examination results, and videos of the drivers under various conditions, including crashes and near-crashes [3]. The former type of information can be anonymized by removing identifying information such as names and replacing them with random dummy variables. However, anonymizing video data is much more challenging, as it requires the removal of difficult identifying information such as facial features while maintaining the scientific value of the data.

Recent developments in computer vision have led to the emergence of Generative Adversarial Networks (GAN)-based techniques for de-identifying face videos [4]. These techniques have the potential to effectively eliminate PII from face videos while preserving human factors cues. In this paper, we aim to explore the effectiveness of GAN-based de-identification techniques for drivers' face videos from NDS data. We provide a comprehensive evaluation of GAN-based de-identification techniques for face videos from NDS data and demonstrate their effectiveness in preserving human factors cues while eliminating PII. For this, we have done extensive experimentation on a large dataset of NDS face videos. We believe this research will contribute to advancing human factors research and improve the responsible sharing of NDS data.

## II. RELATED WORKS

Recently, GANs have been proposed as a promising approach for anonymizing face videos [4]. A GAN is a type of neural network that can generate new data that is similar to a given dataset [5]. In the context of de-identification, GANs can be used to replace the face of a given person with some other face [4]. This is known as "face swapping." One example of a GAN-based face-swapping technique is FSGAN, which is robust in face-swapping and reenactment [6]. Another example is the SF-GAN method, which aims to effectively de-identify faces while preserving important facial attributes like expression, gender, hairstyle, and eyewear [7]. These GAN-based techniques are robust and hence make it difficult to reverse engineer the original faces from the synthetic ones. In addition, they are able to preserve important facial expressions and head movements, making the synthetic faces look more natural and realistic. These GAN-based techniques are promising for anonymizing face videos, as they can effectively conceal the identity of individuals while preserving the visual information of the face. Protecting secondary behavior can be extremely important in transportation research. Thus, GAN-based techniques are a promising area of research for de-identifying face videos in transportation research.

In this paper, we use SimSwap, which is a high-fidelity face-swapping method [8]. It aims to achieve high-fidelity face swapping while preserving facial attributes like expression and gaze direction. The method uses an ID Injection Module, which transfers the identity information of the source face into the target face at the feature level. This allows the algorithm to handle arbitrary face swapping, rather than being limited to specific identities. Additionally, the authors propose Weak Feature Matching Loss, which helps to preserve facial attributes in an implicit way. The authors demonstrate through experiments on wild faces that SimSwap achieves better attribute preservation and competitive identity performance compared to previous state-of-the-art methods. An overview of the use of SimSwap for our purpose is shown in Fig. 1.

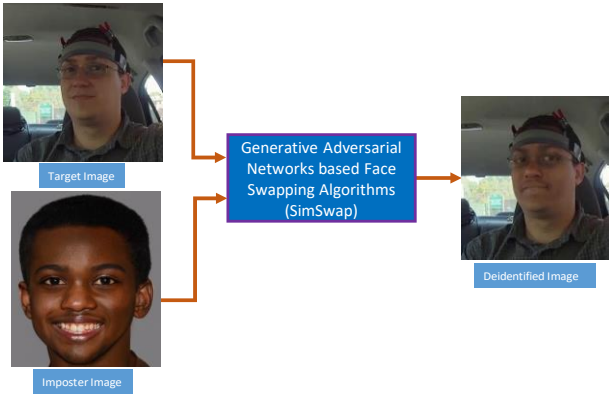

Fig. 1: Overview of face swapping for identification of drivers' face Videos. Interracial combinations can help improve diversity in data for equitable transportation.

## III. DATASET AND HUMAN FACTORS OF INTEREST

The diverse and representative nature of the dataset is essential in accurately evaluating the performance of our proposed method. In this section, we briefly review the dataset and human factors of interest such as head movements, mouth movements, and eye blinking.

### A. Dataset

Assessing the performance of the proposed de-identification framework in naturalistic driving data is crucial for practical applications. The primary source of data used was collected by the Oak Ridge National Laboratory (ORNL)[1]. The data collection was supported by the Exploratory Advanced Research Program of the Federal Highway Administration, with assistance from the Virginia Tech Transportation Institute (VTTI) [9]. The dataset we experimented with had nine participants, who each took a short driving trip lasting between 6 and 10 minutes. The ORNL dataset is rich in terms of the demographic information of the participants. The subjects were a mix of males and females, with a diverse age range, which helped to evaluate the validity of our approach thoroughly.

### B. Behavioral Attributes Necessary in Transportation Research

In the field of transportation research, the study of drivers' behavior through in-cabin videos is crucial. Attributes derived from driver-face videos play an important role in crash analysis and driver attention monitoring. For example, the movements of the eyes can tell a lot about how attentive the driver is while driving the vehicle. Similarly, the mouth movements can tell whether the driver is yawning or not. Apart from that, mouth movements can also indicate if the driver is involved in other secondary activities like eating, drinking, or speaking. Thus, the de-identification of the video should preserve these important features. In this section, we highlight the key behavioral attributes necessary for transportation research derived from drivers' face videos.

*1) Head Pose: Roll, Pitch, and Yaw Angles:* Head pose, which includes roll, pitch, and yaw (RPY) angles, is a crucial factor in driver behavior monitoring [10]. These angles provide information about driver gaze direction and attention. The face is typically considered a rigid body in head pose estimation tasks, with pitch representing up and down movement, yaw representing left and right movement, and roll representing tilt, as shown in Fig. 2. For the calculation of RPY angles, bounding boxes for faces were taken from Retinaface [11] and head pose estimation was done using FSA-Net [12].

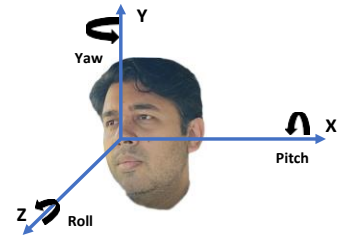

Fig. 2: Head pose angles.

*2) Eye Aspect Ratio:* Eye aspect ratio (EAR) is a value used in computer vision and image processing to determine the state of a person's eyes (open or closed). It is calculated based on the relative positions of various facial features surrounding the eyes and has been used in applications such as drowsy driver detection and facial recognition. The EAR value changes rapidly during the blinking process and hence helps to calculate different eye states. The key points are located using the DLIB library[2] as shown in fig. 3.

---

[1]https://www.ornl.gov/project/ornl-naturalistic-driving-study-sample

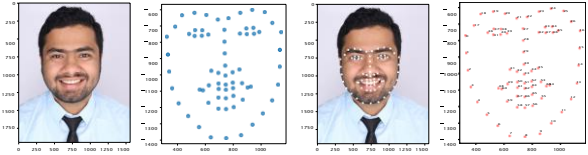

Fig. 3: 68-point facial landmarks given by DLIB.

$$EyeAspectRatio(EAR) = \frac{(d^e_{v1} + d^e_{v2})}{2 \times d^e_h} \quad (1)$$

where, $d^e_{v1}$ is the distance between $P2$ and $P6$ (from Fig. 4). Similarly, $d^e_{v2}$ is the distance between $P3$ and $P5$. $d^e_h$ is the horizontal length of eyes (distance between $P1$ and $P4$).

*3) Pupil Circularity:* Pupil circularity is a measure of the circularity of a pupil in an eye image. It is a value between 0 and 1 that indicates the degree to which the pupil shape deviates from a perfect circle. A value of 1 indicates a perfectly circular pupil, while a value closer to 0 indicates a more elliptical shape. This value can be used to assess eye health, as changes in pupil shape can indicate conditions such as eye fatigue.

$$PupilCircularity = \frac{4 \times \pi \times Area}{Perimeter^2} \quad (2)$$

$Area = (\frac{d^p_r}{2})^2 \times \pi$ ; $d^p_r$ is the distance between P2 and P5.
$Perimeter = d^{p2}_{p1} + d^{p3}_{p2} + d^{p4}_{p3} + d^{p5}_{p4} + d^{p6}_{p5} + d^{p1}_{p6}$ ;
$d^b_a$ is distance between $a$ and $b$ from Fig. 4.

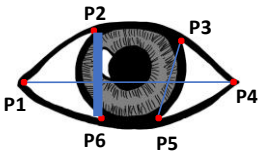

Fig. 4: Landmarks for calculation of EAR.

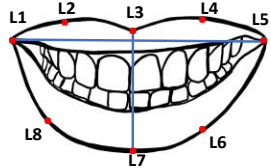

Fig. 5: Landmarks for calculation of LAR.

*4) Lip Aspect Ratio:* Lip aspect ratio (LAR) is a scalar value used to determine the relative size and shape of the upper and lower lips in a face image. It is commonly used in computer vision and face recognition algorithms to identify the position and movements of the lips and to detect various facial expressions such as smiling, frowning, or speaking. The LAR is calculated as the ratio of the distance between the upper and lower lip landmarks divided by the average width of the lips as shown in Fig. 5. The exact formula for calculating LAR may vary based on the specific application and the landmarks used. In our experimentation, LAR is calculated as:

$$LipAspectRatio(LAR) = \frac{d^L_v}{d^L_h} \quad (3)$$

where, $d^L_v$ is the vertical distance between $L3$ and $L7$ (from Fig. 5). Similarly, $d^L_h$ is the horizontal distance between $L1$ and $L5$.

[2]https://github.com/davisking/dlib

Table I shows the statistics of various human cues. It is noteworthy to mention that for EAR and pupil circularity, only subjects without glasses and subjects wearing clear and photochromic glasses were taken.

TABLE I: Statistics of Human Cues for ORNL Dataset

| Human Cues | Maximum | Minimum | Standard Deviation ($\sigma$) |
|---|---|---|---|
| EAR | 0.47 | 0.06 | 0.051 |
| Pupil Circularity | 0.70 | 0.21 | 0.064 |
| LAR | 0.63 | 0.00 | 0.059 |
| Pitch | 45.53 | -54.49 | 8.94 |
| Roll | 45.01 | -38.40 | 6.49 |
| Yaw | 87.77 | -89.19 | 26.27 |

*C. Quantitative Assessment of Image Quality*

Error metrics for assessing image quality are used to calculate the similarity of two images in terms of quality. The metrics used to assess image quality in our experiment are mean squared error (MSE) [13], root mean squared error (RMSE) [14], peak signal-to-noise ratio (PSNR) [15], Universal Image Quality Index (UIQI) [16], Spectral Angle Mapper (SAM) [17], and Relative Dimensionless Global Error Synthesis (ERGAS) [18]. The purpose of using multiple metrics is to increase robustness and account for differences in human perception of image quality. We calculated error metrics for all the de-identified videos vs. original videos. The maximum, minimum, and mean values across all our pairs are given in Table II. The direction of the arrow represents the direction in which images are highly similar. For example, PSNR ↑ means that a higher value of PSNR for a given deidentified and original image pair means they are highly similar.

TABLE II: Error Metric Statistics Across All Frames of ORNL Dataset

| Image Quality Metrics | Original | Deidentified | | |
|---|---|---|---|---|
| | | Minimum | Maximum | Mean |
| MSE ↓ | 0 | 0.873 | 99.39 | 15.88 |
| RMSE ↓ | 0 | 0.935 | 7.969 | 3.909 |
| PSNR ↑ | ∞ | 28.16 | 48.72 | 36.45 |
| UQI ↑ | 1 | 0.439 | 1 | 0.996 |
| ERGAS ↓ | 0 | 247.21 | 16423.93 | 2200.12 |
| SAM ↓ | 0 | 0 | 0.805 | 0.04 |

IV. EXPERIMENTAL SETUP

The experimental setup involved two parts, as shown in Fig. 6: the use of face-swapping algorithms to replace faces in videos, followed by the analysis of the output from the face-swapping algorithms. The videos were first split into frames, and each frame was processed to swap the original face with an imposter's face. In the second part of the experiment, the output from the face-swapping algorithms was analyzed to ensure the deidentification and preservation of human cues important in transportation safety research. Six error metrics were calculated to measure the similarity between the original and imposter faces and ensure that the deidentified face was sufficiently different. The error in human cues (head movement, eye movements, lip movements, etc.) was also calculated and analyzed to ensure that they were similar between the original and imposter faces, preserving important cues for transportation safety research. We also suggest techniques for deidentification of large-scale dataset along with human-in-the-loop validation for possible missed deidentification cases.

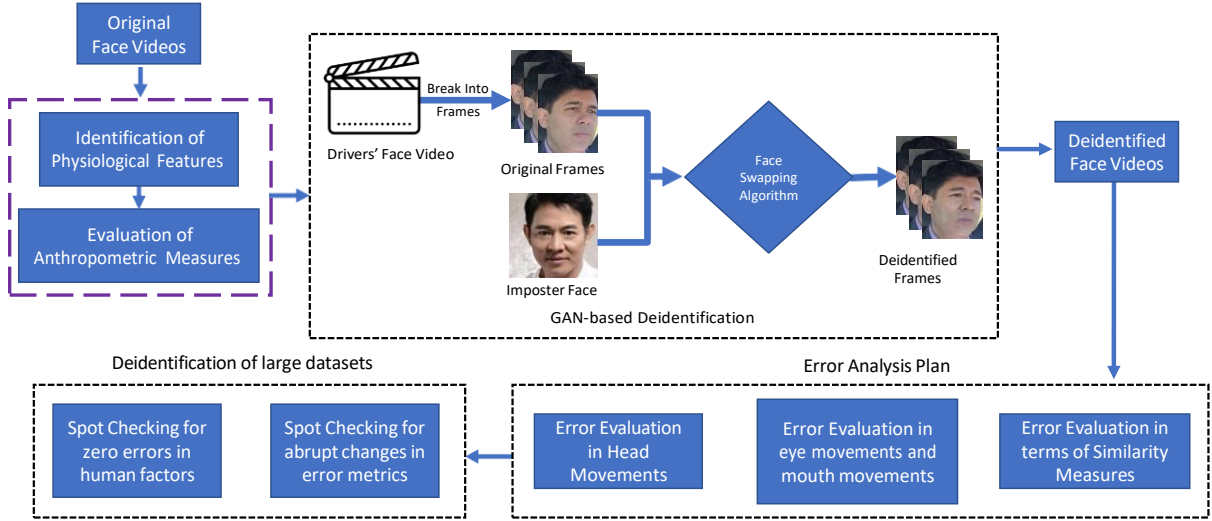

Fig. 6: Overall process for the deidentification of videos along with error analysis plan

## V. Results and Discussion

The error analysis in head movement is reported as errors in RPY angles. Fig. 7 shows some of the results from the ORNL dataset. In order to get an overview across all the head movements, the MAE ($MAE_{rpy}$) for all RPY angles was also calculated.

Fig. 8 shows a side-by-side comparison of the average errors in RPY across the whole ORNL dataset. The yaw error tends to be highly variable, whereas the roll error is the least volatile. The error pattern where Roll Error < Pitch Error < Yaw Error is in fact the expected error scenario. In the naturalistic driving scenario, there is less head-tilt variability, and hence the error in roll is less, as shown in Table I. Head movements up and down are lesser than head movements sideways. For further analysis, we have calculated average errors for different gender combinations for imposter-target pairs which can be seen in Fig. 9. The labels are given as Target-Imposter. For example, the label of "FM" on the X-axis represents that the female face in the video is replaced by a male imposter face. It can be seen that the error is lowest when a female face in the video is replaced by a female imposter face.

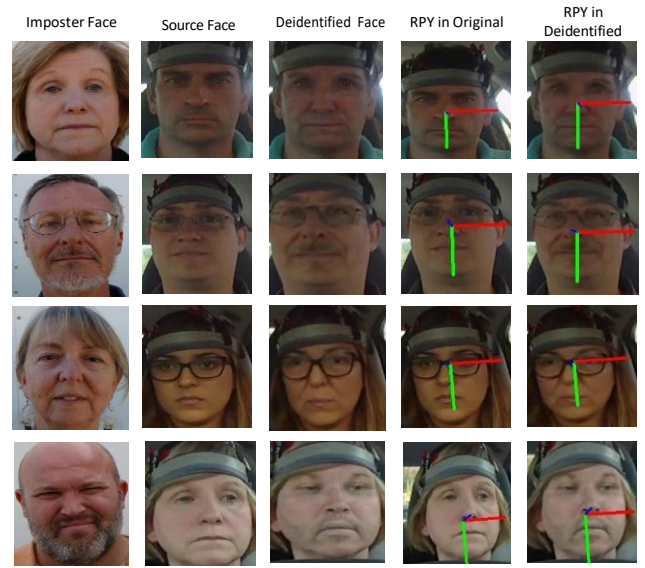

Fig. 7: Results from the de-identification technique. The figures show that the RPY angles are well preserved along with the gaze directions.

Apart from errors in head movements, we also report errors in eye and mouth movements to understand the secondary behavior in de-identified videos. From Fig. 10, the EAR error is less than 0.06 for most of the cases (80% of the cases). The pupil circularity error is also less than 0.075 for most of the cases. This shows that even in the de-identified videos, the EAR and pupil circularity are very well preserved. This error range is acceptable, which shows that the de-identified videos can readily be used in the case of building safe driving models like distracted driving detection models.

### A. Error Analysis Plan

To automate the de-identification of the large datasets using the framework provided, we suggest two major steps.

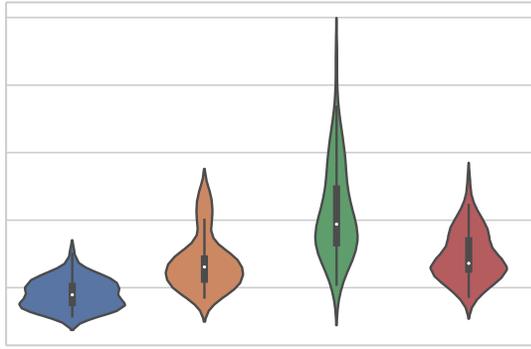

Fig. 8: Violin plot for RPY angular errors along with MAE.

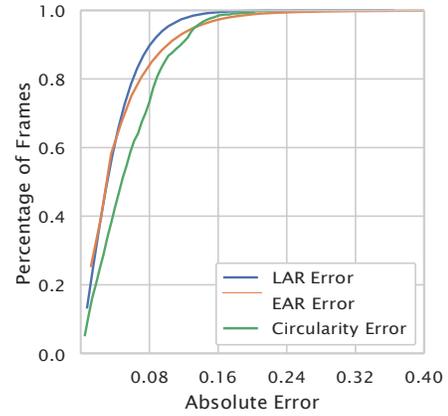

Fig. 10: Percentage of frames vs. absolute error for EAR, LAR, and pupil circularity.

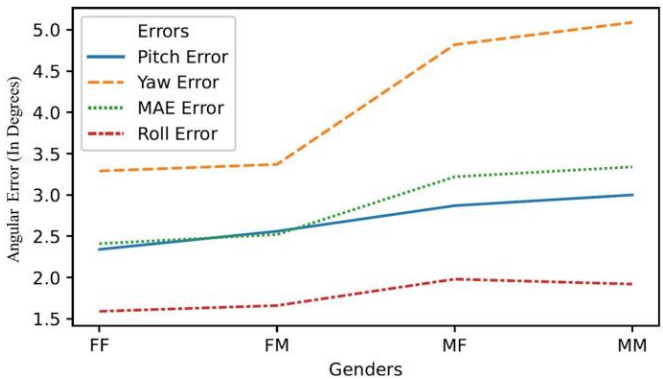

Fig. 9: Error in head movements (in degrees) with respect to gender pairs (Target-Imposter pair).

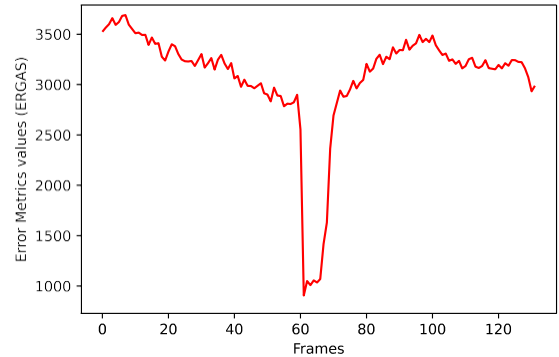

Fig. 11: ERGAS error over the frames of de-identified videos.

**Creating Error Threshold and Spot Checking for Zero Error in Human Factors:** The experimental results showed the need for defining error bounds for various human factors error metrics. This can be done by training on a larger, more diverse dataset. With the ORNL dataset, it was found that for properly de-identified images (visual inspection and not recognized by the recognition algorithm), error (EAR, LAR, RPY, and pupil circularity) was prevalent for all the metrics. Properly de-identified images should not have a zero error across all metrics, and if any single error is outside the acceptable range, it should be subject to manual inspection.

**Spot Checking for Frames with Abrupt Changes in Metrics**: Most quality metrics can detect unusual behavior. As shown in Fig. 11, unusual dips and peaks can be observed. Upon manual inspection, it was found that the given dip was due to improper de-identification in harsh lighting conditions. Human-in-the-loop verification is suggested for such unusual dips and peaks. It is better to take a more sensitive metric like ERGAS (which can range from 0 to tens of thousands).

### B. Testing with NDS Data at VTTI

**Crash Avoidance System Field Operational Test:** The de-identification framework was tested with the Collision Avoidance System Field Operational Test (CAS FOT) face video data of truck drivers collected in an NDS by VTTI. Due to privacy concerns, the data are not shared. The face-swapping algorithms used effectively de-identified the faces while still preserving important attributes related to human factors research, including eye movements, head movements, and mouth movements. The results were evaluated both qualitatively and quantitatively, and the proposed methods were found to be valid. Even with videos taken at night, the de-identification framework worked well.

**1000 Cars NDS Data:** The de-identification framework was tested with a dataset from the 1000 Cars NDS. The results were promising, despite some limitations. The black-and-white video footage did not perform as well as the color video footage, as color information is crucial for facial recognition algorithms to accurately identify and de-identify individuals. Additionally, the framework struggled to effectively de-identify individuals who were wearing glasses, as the eyes appear more bulged in the black-and-white footage and glasses can obstruct important facial features used for identification. Despite these limitations, the framework performed well overall and was able to effectively de-identify a majority of the individuals subjected to testing.

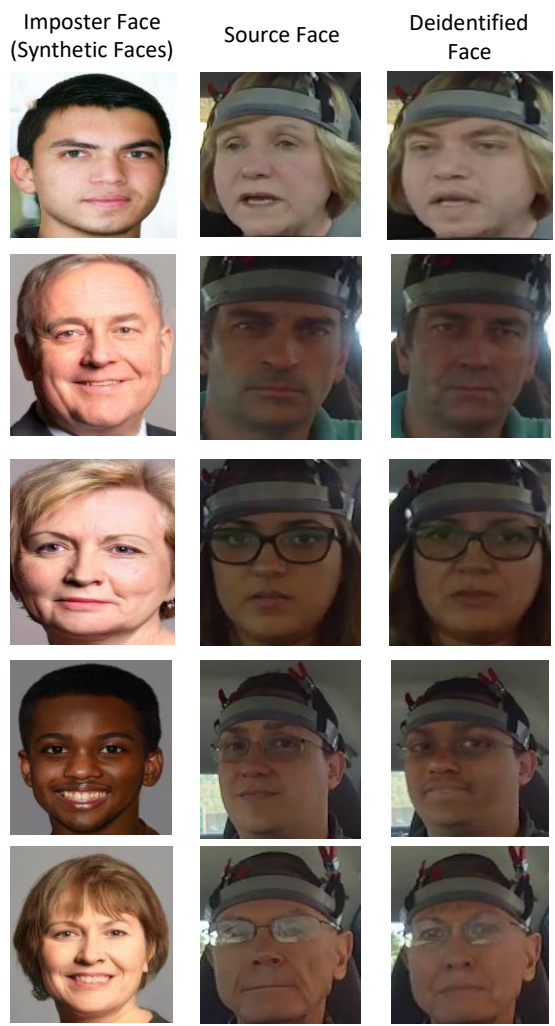

Fig. 12: Use of synthetic faces to replace faces of drivers.

*C. Use of Synthetic Faces in De-identification*

With rising concerns over PII, the use of the faces of real people may again be accompanied by PII concerns. To avoid this, we have used fake faces that do not exist in real life. The faces were imagined by GANs[3] and, hence, such faces can be used to improve diversity in data as well. Fig. 12 shows some examples of when synthetic faces are used for replacing the faces of drivers. It is interesting that the face-swapping techniques hold well even when the participants are wearing glasses.

## VI. CONCLUSION

In this paper, we have presented a way to eliminate data-sharing restrictions due to PII in NDS. We have also proposed ways to quantitatively assess whether the human factors cues necessary for safety research (e.g., head movements, mouth, and eye movements) are preserved. Leveraging the recent advancements in computer vision, this work deals with the removal of biometric identifiers. Although the removal of biometric identifiers is a criterion for the identification of a person, non-biometric identifiers can also be used to identify participants, especially when reidentification attempts are made by people who are already familiar with the participants. For this, future work can explore the possibilities of changing dress, hairstyle, wearables, etc. Also, better threat modeling and more robust human-in-the-loop approaches to spot check the possible cases of failure of de-identification can be explored.

[3]https://vole.wtf/this-mp-does-not-exist/

## REFERENCES

[1] D. L. Longo and J. M. Drazen, "Data sharing," 2016.
[2] B. Fecher, S. Friesike, and M. Hebing, "What drives academic data sharing?," *PloS one*, vol. 10, no. 2, p. e0118053, 2015.
[3] T. A. Dingus, J. M. Hankey, J. F. Antin, S. E. Lee, L. Eichelberger, K. E. Stulce, D. McGraw, M. Perez, and L. Stowe, *Naturalistic driving study: Technical coordination and quality control*. No. SHRP 2 Report S2-S06-RW-1, 2015.
[4] H. Hukkelås, R. Mester, and F. Lindseth, "Deepprivacy: A generative adversarial network for face anonymization," in *Advances in Visual Computing: 14th International Symposium on Visual Computing, ISVC 2019, Lake Tahoe, NV, USA, October 7–9, 2019, Proceedings, Part I 14*, pp. 565–578, Springer, 2019.
[5] I. Goodfellow, J. Pouget-Abadie, M. Mirza, B. Xu, D. Warde-Farley, S. Ozair, A. Courville, and Y. Bengio, "Generative adversarial networks," *Communications of the ACM*, vol. 63, no. 11, pp. 139–144, 2020.
[6] Y. Nirkin, Y. Keller, and T. Hassner, "Fsgan: Subject agnostic face swapping and reenactment," in *Proceedings of the IEEE/CVF international conference on computer vision*, pp. 7184–7193, 2019.
[7] Y. Li, Q. Lu, Q. Tao, X. Zhao, and Y. Yu, "Sf-gan: face de-identification method without losing facial attribute information," *IEEE Signal Processing Letters*, vol. 28, pp. 1345–1349, 2021.
[8] R. Chen, X. Chen, B. Ni, and Y. Ge, "Simswap: An efficient framework for high fidelity face swapping," in *Proceedings of the 28th ACM International Conference on Multimedia*, pp. 2003–2011, 2020.
[9] R. Ferrell, D. Aykac, T. Karnowski, and N. Srinivas, "A publicly available, annotated dataset for naturalistic driving study and computer vision algorithm development," tech. rep., Oak Ridge National Lab.(ORNL), Oak Ridge, TN (United States), 2021.
[10] A. D. Doshi, E. M.-C. Murphy-Chutorian, and M. M. T. Trivedi, "Head pose estimation for driver assistance systems: A robust algorithm and experimental evaluation," 2007.
[11] J. Deng, J. Guo, E. Ververas, I. Kotsia, and S. Zafeiriou, "Retinaface: Single-shot multi-level face localisation in the wild," in *Proceedings of the IEEE/CVF conference on computer vision and pattern recognition*, pp. 5203–5212, 2020.
[12] T.-Y. Yang, Y.-T. Chen, Y.-Y. Lin, and Y.-Y. Chuang, "Fsa-net: Learning fine-grained structure aggregation for head pose estimation from a single image," in *Proceedings of the IEEE/CVF Conference on Computer Vision and Pattern Recognition*, pp. 1087–1096, 2019.
[13] P. Jagalingam and A. V. Hegde, "A review of quality metrics for fused image," *Aquatic Procedia*, vol. 4, pp. 133–142, 2015.
[14] D. Asamoah, E. Ofori, S. Opoku, and J. Danso, "Measuring the performance of image contrast enhancement technique," *International Journal of Computer Applications*, vol. 181, no. 22, pp. 6–13, 2018.
[15] U. Sara, M. Akter, and M. S. Uddin, "Image quality assessment through fsim, ssim, mse and psnr—a comparative study," *Journal of Computer and Communications*, vol. 7, no. 3, pp. 8–18, 2019.
[16] Z. Wang and A. C. Bovik, "A universal image quality index," *IEEE signal processing letters*, vol. 9, no. 3, pp. 81–84, 2002.
[17] O. A. De Carvalho and P. R. Meneses, "Spectral correlation mapper (scm): an improvement on the spectral angle mapper (sam)," in *Summaries of the 9th JPL Airborne Earth Science Workshop, JPL Publication 00-18*, vol. 9, p. 2, JPL publication Pasadena, CA, USA, 2000.
[18] D. Renza, E. Martinez, and A. Arquero, "A new approach to change detection in multispectral images by means of ergas index," *IEEE Geoscience and Remote Sensing Letters*, vol. 10, no. 1, pp. 76–80, 2012.